\documentclass[sigconf,nonacm]{acmart}

\AtBeginDocument{%
  \providecommand\BibTeX{{%
    \normalfont B\kern-0.5em{\scshape i\kern-0.25em b}\kern-0.8em\TeX}}}

\setcopyright{acmcopyright}
\copyrightyear{2021}
\acmYear{2021}
\acmDOI{10.1145/1122445.1122456}

\acmConference[Woodstock '18]{Woodstock '18: ACM Symposium on Neural
  Gaze Detection}{June 03--05, 2018}{Woodstock, NY}
\acmBooktitle{Woodstock '18: ACM Symposium on Neural Gaze Detection,
  June 03--05, 2018, Woodstock, NY}
\acmPrice{15.00}
\acmISBN{978-1-4503-XXXX-X/18/06}

\usepackage{makecell}
\usepackage{hyperref}
\usepackage{graphicx}
\usepackage{amsmath}
\usepackage{natbib}
\usepackage{booktabs} 

\begin{document}
\title{Social Norm Bias: Residual Harms of Fairness-Aware Algorithms}

\author{Myra Cheng}
\email{myra1@stanford.edu}
\affiliation{%
  \institution{California Institute of Technology, Microsoft Research}
  \city{Pasadena}
  \state{California}
  \country{USA}
  }

\author{Maria De-Arteaga}
\affiliation{%
  \institution{University of Texas at Austin}
  \city{Austin}
  \state{Texas}
  \country{USA}
}
\author{Lester Mackey}
\affiliation{%
  \institution{Microsoft Research}
  \city{Cambridge}
  \state{Massachusetts}
  \country{USA}
}
\author{Adam Tauman Kalai}
\affiliation{%
  \institution{Microsoft Research}
  \city{Cambridge}
  \state{Massachusetts}
  \country{USA}
}
\newcommand{\numocc}{25~}
\newcommand{\cov}{$\rho(\textbf{p}_C, \textbf{r}_C)$}
\newcommand{\swe}{\textsc{software}}
\newcommand{\rmsgap}{Gap$^{\textsc{RMS}}$}
\newcommand{\girrev}{$G^{c-\text{irrev}}$}
\newcommand{\rprofnb}{$r^{\textsc{nb}}_{\textsc{professor}}$}
\newcommand{\rprof}{$r_{\textsc{professor}}$}
\newcommand{\snob}{SNoB}
\newcommand{\rirrev}{$r^{\text{irrev}}_c$}
\newcommand{\rirrevbold}{$\textbf{r}^{\text{irrev}}_c$}

\newcommand{\covirrev}{$\rho{(\textbf{p}_C, \textbf{r}^\text{irrev}_C)}$}
\newcommand{\notate}[1]{\textcolor{blue}{\textbf{[LM: #1]}}}
\newcommand{\myranote}[1]{\textcolor{purple}{\textbf{[MC: #1]}}}

\newcommand{\ATK}[1]{\textcolor{red}{\textbf{[ATK: #1]}}}

\renewcommand{\shortauthors}{Cheng et al.}

\begin{abstract}


Many modern machine learning algorithms mitigate bias by enforcing fairness constraints across coarsely-defined groups related to a sensitive attribute like gender or race. However, these algorithms seldom account for within-group heterogeneity and biases that may disproportionately affect some members of a group. 
In this work, we characterize Social Norm Bias (\snob), a subtle but consequential type of algorithmic discrimination that may be exhibited by machine learning models, even when these systems achieve group fairness objectives. We study this issue through the lens of gender bias in occupation classification. We quantify \snob~by measuring how an algorithm's predictions are associated with conformity to inferred gender norms. When predicting if an individual belongs to a male-dominated occupation, this framework reveals that ``fair'' classifiers still favor biographies written in ways that align with inferred masculine norms. We compare \snob~across algorithmic fairness methods and show that it is frequently a residual bias, and post-processing approaches do not mitigate this type of bias at all.
\end{abstract}  
\maketitle

\section{Introduction}

As automated decision-making systems play a growing role in our daily lives, concerns about algorithmic unfairness have come to light. It is well-documented that machine learning (ML) algorithms can perpetuate existing social biases 
\citep{buolamwini2018gender, noble2018algorithms, stark_stanhaus_anthony_2020}. To avoid algorithmic discrimination based on sensitive attributes, various approaches to measure and achieve ``algorithmic fairness'' have been proposed. These approaches are typically based on \textit{group fairness}, which partitions a population into groups based on a protected attribute (e.g., gender, race, or religion) and then aims to equalize some metric of the system across the groups \citep{hardt2016equality, agarwal2018reductions,dwork2012fairness, pleiss2017fairness, kamiran2012decision,lohia2019bias}.

Group fairness makes the implicit assumption that a group is defined solely by the possession of particular characteristics \citep{hu2020s}, ignoring the heterogeneity within groups. For example, most work on gender bias in ML compares groups (and typically assumes there are only two groups); however, different women's experiences of gender and concept of womanhood vary drastically \citep{shields2008gender,wood2009gender}.
These approaches do not account for the complex, multi-dimensional nature of concepts
like gender and race
\citep{hanna2020towards, butler2011gender}, and how this intersects with discrimination. 
Consider, for instance, social norms which represent the implicit expectations of how particular groups behave. In many settings, adherence to or deviations from social norms result in harm, but current algorithmic fairness approaches overlook whether and how ML replicates these harms. 
 
Social norms may influence an algorithm's predictions even after group fairness approaches have been applied. We characterize Social Norm Bias (\snob)\textemdash the associations between an algorithm's predictions and individuals' adherence to inferred social norms\textemdash as a source of algorithmic unfairness. We propose an approach to measure inferred social norms, and show that the inferred norms, specific to a dataset, reflect real-world norms and stereotypes that arise in certain settings. Thus, penalizing individuals for their adherence or deviation to social norms is an overlooked form of algorithmic bias and a residual harm of many fairness-aware algorithms.  

We study \snob~through the task of occupation classification, a prediction problem relevant in various applications ranging from targeted job opportunity advertisements~\citep{de2019bias} to Google's ``People Cards.'' The risk of (group) gender bias in this task has been studied in the literature~\citep{de2019bias,ceren24gender}, and Twitter users have expressed concern over Google search results mislabeling famous women scientists as teachers while labeling their male peers correctly.\footnote{Twitter thread started by Dr. Timothy Verstynen:
 https://twitter.com/tdverstynen/status/1501386481415434245}  
In this setting, \snob~occurs when an algorithm favors biographies written in ways that adhere to inferred gender norms of the majority, which we show reproduce real-world norms. These associations, which remain even after applying bias-mitigation approaches, may lead to representational and allocational harms for feminine-expressing people in male-dominated occupations 
\citep{bartl2020unmasking, blodgett2020language}. For example, consider a male-dominated occupation in which men are less likely to present themselves as family-oriented, or more likely to describe themselves using adjectives that denote success \citep{garg2018word}. \snob~occurs when an algorithm is more likely to correctly classify the women in this occupation who adhere to these norms over the women in the same occupation who do not adhere to them. This phenomenon compounds existing patterns of discrimination and gender bias in the workplace.

Our framework quantifies algorithmic bias on a dimension that extends beyond group-level disparities, illuminating a gap between algorithmic harms and current techniques for algorithmic fairness. We find that approaches aiming to improve group fairness, which we refer to as fairness-aware classifiers, may still exhibit \snob. In particular, post-processing approaches that maintain within-group ordering are especially prone to producing harms on the basis of inferred social norms, while some in-processing approaches mitigate this type of bias. Failing to explicitly account for and assess \snob~may lead to the deployment of ``fair'' algorithms that not only perpetuate these harms but also obscure them.
 
\section{Related Work}\label{relwork}
It is well-established that ML models can reflect and amplify societal biases. For example, semantic representations such as word embeddings encode gendered and racial biases \citep{bolukbasi2016man, caliskan2017semantics, rudinger-etal-2017-social, swinger2019biases}. In coreference resolution, systems associate and thus assign gender pronouns to occupations when they co-occur more frequently in the data \citep{rudinger2018gender}.
In toxicity detection, language related to marginalized groups is disproportionately flagged as toxic \citep{zhou2021challenges}. Various fairness-aware algorithms aim to mitigate these issues from a group fairness perspective
\citep{park-etal-2018-reducing, kumar-etal-2020-nurse,bordia-bowman-2019-identifying, gonen2019lipstick}.

However, algorithms that satisfy group fairness metrics may still exhibit discriminatory behavior and cause harm to marginalized populations based on \textit{reductive definitions} \citep{antoniak-mimno-2021-bad}. Our work identifies a concrete type of harm, arising from reductive definitions, that is not captured by group-based metrics of bias and frequently remains in fairness-aware algorithms. We go beyond group fairness to characterize algorithmic predictions' associations with inferred social norms. Our approach builds upon work such as \citet{cryan2020detecting} and \citet{tang2017gender} to assess inferred gender norms. \citet{cryan2020detecting} develop an ML-based method to detect gender stereotypes and compares the approach to crowd-sourced lexicons. \citet{tang2017gender} take a simpler approach, relying on manually-compiled lists of gendered words and using only the frequency of these words as a measure of gender bias.

There is a growing body of work on how algorithms operationalize stereotypes, which are related to social norms \citep{nadeem2020stereoset, nangia2020crows,blodgett-etal-2021-stereotyping}. 
We focus on social norms as a broader phenomenon than stereotypes \citep{russell2012perceptions}. Since  inferred social norms arise from the structure of the dataset, we capture not only stereotypes but gender-related patterns present in the data. For example, the mention of a women's college or fraternity in one's biography is a proxy for gender; it is a norm rather than a stereotype that women's colleges are attended by women while members of fraternities are men.

Adherence to social norms is a form of hetereogeneity within groups, which is connected to the concept of subgroup fairness. For example, work on multicalibration \citep{hebert2018multicalibration} and fairness gerrymandering \citep{kearns2017preventing}
provide theoretical perspectives on bridging the gap between group and individual fairness by mitigating unfairness for subgroups within groups, even when they are not explicitly labeled. Our approach differs from this work by focusing on a specific type of harm that affects subgroups, drawing its connection to social discrimination and providing an approach to assess it in practice.

Algorithmic fairness definitions that extend beyond group disparities and capture causal dependencies, such as counterfactual fairness~\citep{kusner2017counterfactual}, are in part motivated by concerns similar to those underlying \snob. Accounting for causal dependencies between group membership and other features would help account for social norms. However, causal graphs are rarely available. Our work proposes an approach to capture (some) dependencies using a statistical machine learning model, rather than a causal model, which enables the assessment in practice of an important type of residual harms.

\section{Background}\label{bg}
Considering sensitive group membership alone is not sufficient to establish fairness. While the relevance of \snob~is not exclusive to the task of occupation classification nor to gender, its potential harms stem from tight connections with existing patterns and forms of discrimination in contexts where individuals that belong to a marginalized group but have more similarities to the dominant group, such as adherence to social norms, are treated more favorably. Thus, in this paper we study \snob~through a case study in the occupation classification context. In this section, we provide background on various axes of gender and how they are operationalized in the workplace and in the use of English language. We also describe existing approaches to address gender bias in tasks related to automated hiring.

\paragraph{The Multiplicity of Gender}\label{multiplicity}
The term ``gender'' is a proxy for different constructs depending on the context \citep{keyes2021you}. It may mean \textit{gender identity}, which is one's “felt, desired or intended identity” \cite[p.2]{glick2018data}, or 
\textit{gender expression}, which is how one ``publicly expresses or presents their gender... others perceive a person’s gender through these attributes'' \cite[p.1]{ontario}. These concepts are also related to \textit{gender norms}, i.e. ``the standards and expectations to which women and men generally conform,'' \cite[p.717]{mangheni2019gender} including personality traits, behaviors, occupations, and appearance \citep{agius2012trans}. 
These various notions of social gender encompass much more than the categorical gender labels that are used as the basis for group fairness approaches \citep{cao2019coref}. Using name or pronouns as a proxy for gender \citep{sen2016race, bolukbasi2016man} reinforces categorical definitions, which leads to further harms to those who are marginalized by these reductive notions \citep{keyes2021you,larson2017gender,mitchell2020diversity}. We focus on discrimination related to the ways that individuals' gender expression adhere to social gender norms.

\paragraph{Harms Related to Gender Norms in the Workplace}\label{genderlit}
Our concerns about the use of gender norms in ML systems are grounded in studies of how gender norms have been operationalized in various occupations, causing harm to gender minorities. 
It is well-established that ``occupations are socially and culturally `gendered'''
\citep{stark_stanhaus_anthony_2020}; many jobs in science, technology, and engineering are traditionally masculine \citep{ensmenger2015beards,light1999computers}.
In social psychology, \textit{descriptive stereotypes} are attributes believed to characterize women as a group. The perceived lack of fit between feminine stereotypic attributes and male gender-typed jobs result in gender bias and impedes women's careers \citep{heilman2001description, heilman2012gender}. If and when these patterns are replicated by ML algorithms, this results in two types of harms. The associations that we highlight may lead to 1) representational harm, when actual members of the occupation are made invisible, and 2) allocational harm, when certain individuals are allocated fewer career opportunities based on their degree of adherence to gender norms \citep{bartl2020unmasking, blodgett2020language}.

\paragraph{Gender Norms in Language}\label{genderclasser}
Different English words have gender-related connotations \citep{moon2014gorgeous}. Crawford et al. (\citeyear{crawford_leynes_mayhorn_bink_2004}) provide a corpus of 600 words and human-labeled gender scores, as scored on a 1-5 scale (1 is the most feminine, 5 is the most masculine) by undergraduates at U.S. universities. They find that words referring to explicitly gendered roles such as \textit{wife, husband, princess,} and \textit{prince} are the most strongly gendered, while words such as \textit{pretty} and \textit{handsome} also skew strongly in the feminine and masculine directions, respectively.
Gender also affects the ways that people are perceived and described \citep{madera2009gender}.
A study on resumes describes gender-based disparities in how people represent themselves, despite having similar experiences \citep{snyder2015resume}. 
In letters of recommendation, women are described as more communal and less agentic than men, which harms the women who are perceived in this way \citep{madera2009gender}. 
Wikipedia articles also contain significant gender differences, such as notable women's biographies focusing more on their romantic and family lives \citep{wagner2015s}.

\paragraph{Addressing Gender Bias in Automated Hiring}
Audit studies reveal that employers tend to discriminate against women \citep{bertrand2004emily,johnson2016if}. 
These biases are also replicated in automated hiring, as demonstrated by the gender gap in error rates of an occupation classification algorithm \citep{de2019bias}. Many in academia and industry alike have been motivated to mitigate these concerns \citep{raghavan2020mitigating,bogen2018help,sanchez2020does}.
For instance, LinkedIn developed a post-processing approach for ranking candidates so that their candidate recommendations are demographically representative of the broader candidate pool; their system is deployed across a service affecting more than 600 million users \citep{geyik2019fairness}. Other intervention techniques have also been proposed \citep{dwork2018decoupled, romanov2019s}. These approaches share a reliance on categorical gender labels to measure fairness.

\section{Measuring Social Norm Bias}\label{meth}

We develop a framework to measure \snob~in a classification model by 1) identifying inferred social norms specific to a dataset and 2) measuring the correspondence between these norms and the outcomes of the model. 

\subsection{Identifying Inferred Social Norms}\label{inferrednorms}
We use a supervised learning approach to infer social norms from a dataset. We train a predictive model $G$ on the dataset to distinguish between data labeled with different values of $l$, where $l$ is the group label (corresponding to a sensitive attribute) that takes on $n$ categorical values. 

For a given datum $x$, let $G_{l}(x)$ be the predicted probability that $x \in L_l$, where $L_l$ is the subset of the data that has group label $l$. We refer to $G$ as a norm scorer, as its output is indicative of how much $x$ adheres to a set of norms, i.e. dominant characteristics, of $L_l$. For example, to study inferred gender norms in a dataset of biographies, the gender label $L$ is obtained from the pronouns that someone uses in their biography. $G_{\text{she}}(x)$ is the predicted probability that $x$ uses ``she'', which is a measure of adherence to feminine social norms inferred from the dataset. Due to predictive multiplicity~\citep{marx2020predictive}, there is not a unique norm scorer, thus, \snob~does not constitute a unique metric inherent to a dataset and classifier pair, but rather a diagnostic framework to identify potential residual harms. In our experiments, we show results for two variants of a norm scorer, one of which also integrates task irrelevance constraints.  

\begin{table*}[t]
  \caption{Notation used throughout the paper.  \cov, \rmsgap, and \covirrev~are detailed in Sections  \ref{covdef}, \ref{rmsexp}, and \ref{robust}, respectively. } 
  \label{tab:notation}
\begin{small}

\begin{tabular}{lp{0.74\textwidth}}
\toprule
\textbf{Term}& \textbf{Definition}\\
\midrule	
$c$ & occupation \\
$C$ & set of all occupations \\
$H_c \in [0,1]$ & set of biographies in $c$ using ``he'' \\
$S_c \in [0,1]$ & set of biographies in $c$ using ``she'' \\\midrule
$p_c \in [0,1]$ &  fraction of ``she" biographies in $c$, measure of gender imbalance in $c$\\
$r_c \in [0,1]$ & correlation between target and gender norm scores across $S_c$\\
$\textbf{p}_C \in [0, 1]^{|C|}$ & $\{p_c|c \in C\} $\\
$\textbf{r}_C \in [0, 1]^{|C|}$ & $\{r_c|c\in C\} $\\
\cov~$\in [0,1]$ & correlation between $\textbf{p}_C$ and $\textbf{r}_C$, \snob~metric for a classifier\\\midrule
\rmsgap & gender gap in TPR across $C$, group fairness metric for a classifier\\
\midrule
\rirrev~$\in [0,1]$ & correlation between target and occupation-irrelevant gender norm scores across $S_c$\\
\rirrevbold~$\in [0, 1]^{|C|}$ & $\{$\rirrev$|c\in C\} $\\
\covirrev$\in [0,1]$ & correlation between $\textbf{p}_C$ and \rirrevbold, robustness check for \cov\\
\bottomrule

\end{tabular}		

\end{small}

\end{table*}

\subsection{Measuring Social Norm Bias}\label{rcscore}
\paragraph{Binary Classification Setting}\label{singleocc} Assume that a binary classification algorithm outputs the predicted probability $\hat{Y}(x)$ that the label of the datum $x$ is $Y$. To evaluate \snob, we consider the correlation $r$ between $G_l(x)$ and $\hat{Y}(x)$, the predicted probabilities from the two models, across the subset $L_l$. Specifically, we compute Spearman's rank correlation coefficient $r$ between $\{\hat{Y}(x) | x\in L_l,  Y(x) = 1\}$ and $\{G_l(x) | x \in L_l,  Y(x) = 1\}$. We use Spearman's correlation rather than Pearson's since the latter measures only the strength of a linear association, while we aim to capture general monotonic relationships.  The magnitude of $r$ is a measure of the degree of \snob~exhibited by the algorithm $\hat{Y}$ with respect to norms $G_l$. Since $r$ is computed only from the subset $L_l$, this metric captures associations present \textit{within} the group; that is, it establishes if within-group ordering of predicted probabilities within a class is correlated with adherence to inferred norms. 

\paragraph{Multiclass Classification Setting}\label{covdef}
In the multi-class classification problem, where $C$ is the set of classes, the algorithm outputs the predicted probability $\hat{Y}_c(x)$ that the label of the datum $x$ is $c$, for $c\in C$. 
Let $$p_c = \frac{|\{x|x\in L_l, Y_c(x) = 1\}|}{|\{x |Y_c(x) = 1\}|},$$ i.e. the fraction of data in class
$c$ that have group label $l$. Let $r_c$ be the metric $r$ for a specific class $c$, i.e. the correlation between $\{\hat{Y}(x) | x\in L_l,  Y_c(x) = 1\}$ and $\{G_l(x) | x \in L_l,  Y_c(x) = 1\}.$
To quantify \snob~across all classes, we use 
\cov, the Spearman's rank correlation coefficient between $\textbf{p}_C$ and $\textbf{r}_C$, where $$\textbf{r}_C = \{r_c|c\in C\} \in [0, 1]^{|C|},$$$$  \textbf{p}_C = \{p_c|c \in C\} \in [0, 1]^{|C|}.$$ The quantity \cov~enables us to compare the extent to which an algorithm relies on inferred social norms across different classes, and whether this is correlated with the group imbalance across classes. For example, in the occupation classification setting, gender imbalance varies by occupation, and \cov~establishes whether the likelihood of correctly predicting that someone belongs to a male-dominated occupation has higher associations with masculine gender norms as gender imbalance in the occupation increases.

\section{Inferred Social Norms in Occupation Classification}

We present a case study of \snob~in occupation classification, a task relevant to various real-world recruiting and hiring applications, and in which \snob~may compound harms. Several forms of online presence, such as a personal websites, often do not list individuals' occupations in a structured way. Thus, ML methods to automatically classify occupation can support companies conducting targeted recruiting efforts \citep{de2019bias,peng_investigations_2022}. We assume that a fair occupation classification algorithm should not exhibit gender bias, including any association with inferred social norms, since career prospects should not depend on gender. 

\subsection{Occupation Classification}\label{occtask}
\paragraph{Dataset}
We use the dataset\footnote{The dataset is publicly available at \href{http://aka.ms/biasbios}{http://aka.ms/biasbios} and licensed under the MIT License.} and task described by \citet{de2019bias}. 
The dataset, containing 397,340 biographies spanning twenty-eight occupations, is obtained by using regular expressions to filter the Common Crawl for online biographies. Each biography is labeled with its gender based on the use of ``she'' or ``he'' pronouns (in Appendix \ref{nbanal}, we study \snob~on a newly-collected small set of biographies with nonbinary pronouns).
Let $H_c$, $S_c$ be the sets of biographies in occupation $c$ using ``he'' and ``she'', respectively. $|H_c|,|S_c|$ are the numbers of biographies in the respective sets. To preserve the ratios between $|H_c|$ and $|S_c|$, we use a stratified split to create the training, validation, and test datasets, containing 65\%, 10\%, and 25\% of the biographies respectively. This split is consistent with other work that uses this dataset \citep{de2019bias, romanov2019s}.
We use the data to train and evaluate algorithms that predict a biography's occupation title from the subsequent sentences. For training, we remove names, titles, and explicit gender indicators.\footnote{Consistent with previous work \citep{de2019bias}, we used regular expressions to remove the following words from the data: he, she, her, his, him, hers, himself, herself, mr, mrs, ms, ph, dr. }

\paragraph{Semantic Representations and Learning Algorithms}\label{models}
For the occupation classification algorithms, we use three semantic representations with different degrees of complexity: bag-of-words, word embeddings, and BERT.
In the bag-of-words (BOW) representation, a biography $x$ is represented as a sparse vector of the frequencies of the words in $x$. 
BOW is widely used in settings where interpretability is important.
In the word embedding (WE) representation, $x$ is represented by an average of the fastText word embeddings \citep{bojanowski2017enriching, mikolov2018advances} for the words in $x$. 
Previous work demonstrates that the WE representation captures semantic information  effectively \citep{adi2016fine}. For the BOW and WE representations, we train a one-versus-all logistic regression model with $L_2$ regularization on the training set, as done by \citet{de2019bias}. 
 The BERT contextualized word embedding model \citep{devlin2018bert} is state-of-the-art for various natural language processing tasks, and it has been widely adopted for many uses. Unlike the other language representations, a biography's BERT encoding is context-dependent. The BERT representation of a word depends on its meaning within the sentence, which is determined by an attention mechanism that considers the surrounding words in the word's sentence. We fine-tune the BERT model, which pre-trains deep bidirectional representations from unlabeled English text \citep{wolf-etal-2020-transformers}, for the occupation classification task.
 
\paragraph{Group Fairness Metric: \rmsgap}\label{rmsexp}
Prior research on group fairness in occupation classification \citep{de2019bias, romanov2019s}  focus on reducing \rmsgap, a measure of the difference in the classifier's true positive rate (TPR) between ``she'' and ``he'' data. Let $a, \neg a$ be values of the group label for the sensitive attribute $l$ (gender).
For an occupation $c \in C$,  we have
$$\textsc{TPR}_{c,a} = \Pr[\hat Y_c = 1 | Y_c = 1, l = a], $$$$\textsc{Gap}_{c,a} = \textsc{TPR}_{c,a} - \textsc{TPR}_{c,\neg a}.$$
Then, \rmsgap~is defined as:
 $$\textrm{Gap}^{\textsc{RMS}} = \sqrt{\frac{1}{|C|}\sum_{c\in C} \textsc{Gap}^2_{c,a}}.$$
Root mean square (RMS) penalizes larger values of the gender gap more heavily than a linear average across the occupations. This metric is closely related to equality of opportunity, a measure of fairness introduced by \citet{hardt2016equality}.

\subsection{Inferred Gender Norms}\label{langgender}
We apply the method described in Section \ref{inferrednorms} to infer gendered social norms specific to the dataset and obtain a score of how each individual adheres to the norms of ``s/he'' biographies using a logistic regression model. In this setting, the norms arise from the the natural language properties of our data. As discussed in Section \ref{genderclasser},  language is deeply gendered.
Since some occupations are dominated by biographies of a single gender, we pre-process the data such that the ``she'' and ``he'' biographies in each occupation are weighted equally, preventing $G$ from learning to identify gender from occupation alone \citep{wang2019balanced}. Specifically, we assign weight $\alpha_x$ to a biography $x$ as follows: 
$$\alpha_x =
\begin{cases}
 \frac{|H_c|}{\text{max}\{|S_c|, |H_c|\}} & \quad \text{if}~x \in S_c, \\
 \frac{|S_c|}{\text{max}\{|S_c|, |H_c|\}} & \quad  \text{if}~x \in H_c.
\end{cases}$$

We train $G$ using the WE semantic representation. $G$ achieves 0.68 accuracy. In Section \ref{realworld}, we show correspondence between these inferred norms and real-world gender norms.

\section{Social Norm Bias in Occupation Classification}\label{compfairsection}
Algorithmic group fairness approaches fall broadly into three paradigms: pre-, post-, and in-processing techniques \citep{kamiran2012data}. We describe the algorithmic fairness approaches that we evaluate, and then compare their \snob~in the occupation classification setting.

\subsection{Fairness Intervention Techniques}
\textbf{Pre-processing} fairness approaches modify the data before it is used to train the algorithm, such as by reweighting the training distribution to account for representation imbalances or by changing features of the data \citep{calmon2017optimized,kamiran2012data}. Since many of the occupations are gender-imbalanced in our dataset, we adopt the pre-processing approach of re-weighting the data such that half of the total weight belongs to each of $S_c$ and $H_c$ (set of biographies in c using ``she" and ``he", respectively), using the weights $\alpha_x$ as described in Section \ref{langgender}.

\textbf{Post-processing} fairness approaches
apply an intervention after training the algorithm to balance some metric across groups \citep{pleiss2017fairness, kamiran2012decision,lohia2019bias,hardt2016equality}. One common approach (PO) is to change the decision threshold for particular groups to equalize relevant fairness-related metrics, such as false positive rates or acceptance rates \citep{hardt2016equality}. These approaches are relatively cost-effective to implement since they do not require re-training and have been deployed in large-scale systems such as on LinkedIn \citep{geyik2019fairness}.

\textbf{In-processing} group fairness approaches modify the algorithm at training time. We evaluate a range of state-of-the-art in-processing approaches, including decoupled classifiers, reductions, Covariance Constrained Loss (CoCL), and adversarial learning. The decoupled approach is relatively simple, with a separate classifier trained for each gender group \citep{dwork2018decoupled}. 

The reductions method is the primary in-processing mitigation method in the Fairlearn Python package \citep{bird2020fairlearn}. It reduces a classification task to a sequence of cost-sensitive classification problems \citep{agarwal2018reductions}. We consider it only for the WE representation since it is too computationally expensive to use with BERT and unable to reduce either \rmsgap~or \cov~with BOW. 

We evaluate CoCL, which differs from most other in-processing techniques in that it does not require access to gender labels. The method adds a penalty to the loss function that minimizes the covariance between the probability that an individual's occupation is correctly predicted and the word embedding of their name~\citep{romanov2019s}. 
\citet{romanov2019s} validate CoCL's effectiveness in reducing \rmsgap~on the same biographies dataset using the BOW representation, which we also use. 

Finally, the adversarial learning method optimizes directly for independence between sensitive group membership and predictions, while accounting for dependencies between features. It learns to make predictions that maximize accuracy while reducing an adversary's ability to determine the protected attribute from the predictions \citep{zhang2018mitigating}. We use the implementation in the AI Fairness 360 Python package \citep{bellamy2019ai}.
 
 \subsection{Measuring \snob}
To measure \snob~in the occupation classification setting, we use the metrics $r_c$ and \cov~introduced in Section \ref{rcscore}. We consider the subset with group label ``she'', i.e. we use $S_c$ as $L_l$. For each occupation $c$, we compute the correlation $r_c$ between $\{\hat{Y}_c(x) | x\in S_c\}$ and $\{G(x) | x \in S_c\}$, i.e. across the ``she'' biographies in the occupation. A positive/negative value indicates that adherence to more feminine/masculine norms are rewarded by $\hat{Y}_c$. 

The fraction of biographies in occupation $c$ that use ``she'' is  $$p_c = \frac{|S_c|}{|S_c|+|H_c|}.$$  If $p_c < 0.5$, $c$ is male-dominated, and vice versa. If $r_c$ is more negative in more male-dominated occupations, this implies that individuals whose biographies are more aligned to masculine gender norms are also more likely to have their occupation correctly predicted by the occupation classification algorithm. 

\cov~quantifies the relationship between gender imbalance and $r_c$ across the set of all occupations $C$. A high value of \cov~implies that in more gender-imbalanced occupations, $r_c$ is larger in magnitude. Furthermore, a positive \cov~indicates that the more gender-imbalanced an occupation is, the more it favors adherence to gender norms of the over-represented gender. This notation is summarized in Table \ref{tab:notation}.

\begin{figure}[h]
  
  \centering
  \includegraphics[width=\linewidth]{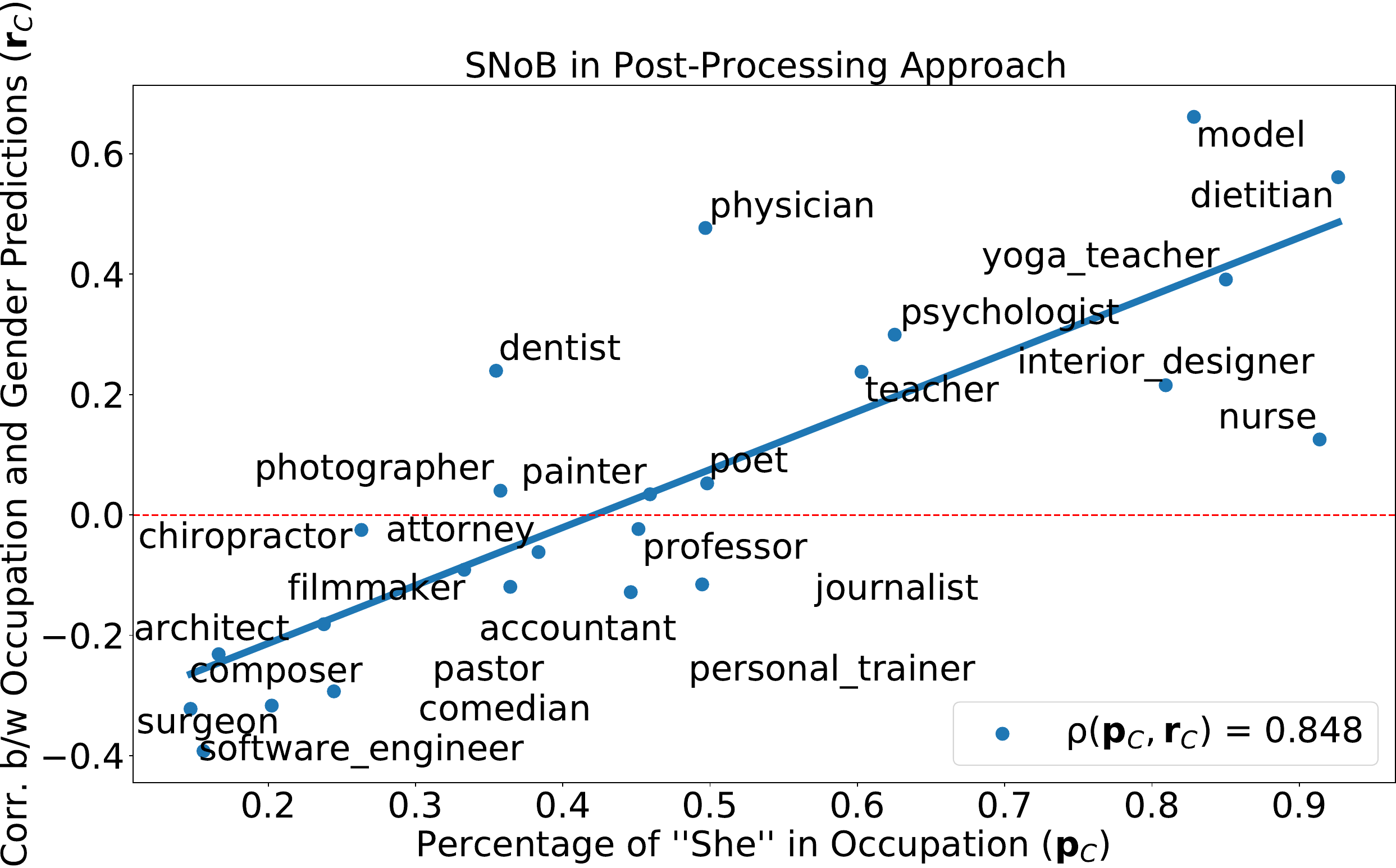}
  
  \caption{\textbf{\snob~across occupations using a post-processing approach.} 
  The extent to which an algorithm's predictions align with gender norms ($y$-axis) is correlated with the gender imbalance in the occupation ($x$-axis). Ideally, without any \snob, the correlation $r_c = 0$, so every point would lie on the dotted red line. The WE representation is depicted here, and other representations (BOW, BERT) have similar trends. Note that these values are the same for the fairness-unaware approach as the post-processing approach, since the latter does not change the within-group ordering of individuals. 
  }
    \label{fig:ppcorr}
    
\end{figure}
\begin{figure}[h]
  \centering
  \includegraphics[width=\linewidth]{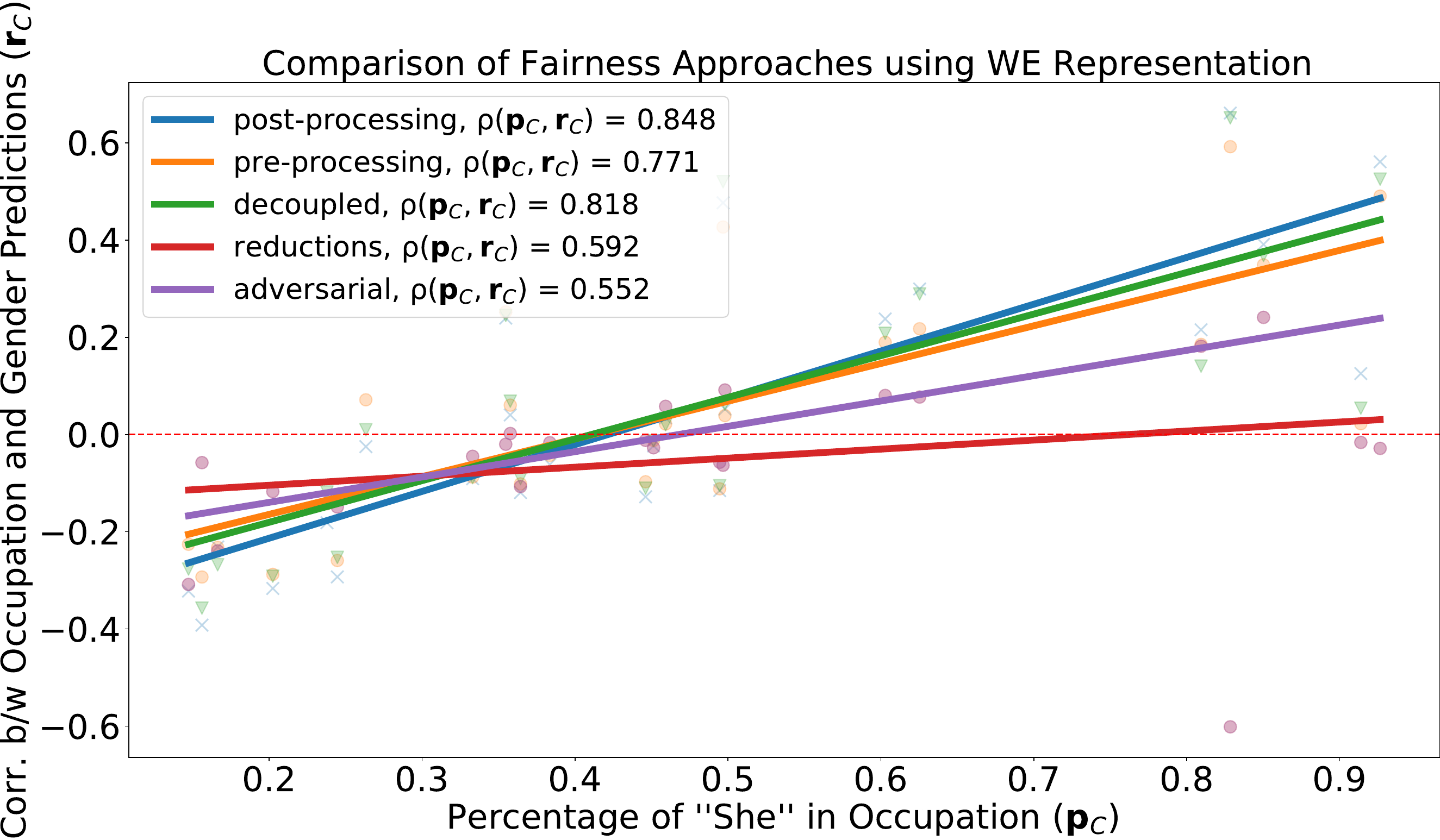}
  \caption{\textbf{Comparing fairness interventions.}  While \snob~persists across group fairness interventions, it is somewhat mitigated by the in-processing approaches. It is minimized by the adversarial technique.
  }
    \label{fig:approaches}
    
\end{figure}

\subsection{\snob~vs. Group Fairness Metrics for Different Approaches}\label{diff}
We use \cov~to compare the cross-occupation associations of different fairness approaches. Table \ref{tab:faircompare} lists the accuracy, group fairness metric \rmsgap, and \cov~for each approach. Figure \ref{fig:ppcorr} plots the gender proportion in each occupation versus its value of $r_c$ when using the post-processing approach with a WE representation, and Figure \ref{fig:approaches} visualizes this trend for various fairness-aware classifiers. 

\begin{table}[t]
  \caption{Comparison of group fairness and \snob~metrics across fairness approaches.
  Refer to Table \ref{tab:notation} for definitions of
  \cov, \rmsgap, and \covirrev~, which are introduced in Sections \ref{covdef}, \ref{rmsexp}, and \ref{robust}, respectively. 
  For BOW and WE, the one-versus-all $\hat{Y}_c$ accuracy is averaged across all occupations. For BERT, the model is a multi-class classifier. Although post-processing fairness intervention techniques mitigate \rmsgap~the most, they have the highest values of \cov. This suggests that they are the least effective at reducing \snob.\protect\footnotemark{}} 
  \label{tab:faircompare}
\begin{center}
\begin{small}
\begin{tabular}{lcccccr}
\toprule
Approach&$\hat{Y}_c$ Acc&\rmsgap&\cov&\covirrev\\
\midrule	
BOW, pre-process & $0.92$ & $0.078$ & $0.79^{**}$ & $0.44^{*}$\\
  BOW, post-process & $0.95$ & $0$ & $0.80^{**}$ &$0.52^{**}$\\
   BOW, decoupled &$0.96$ & $0.10$ & $0.78^{**}$ & $0.26$\\
    BOW, CoCL\footnotemark{} & $0.95$ & $0.074$ & $0.74^{**}$ & $0.46$\\
    BOW, adversarial & $0.98$ & $0.14$ & $0.41^{*}$ & $0.044$\\\midrule
   WE, pre-process & $0.94$ & $0.061$ & $0.77^{**}$ & $0.32 $\\
    WE, post-process & $0.97$ & $0$ & $0.85^{**}$ & $0.60^{**}$\\
    WE, decoupled & $0.94$&$0.060$ & $0.82^{**}$ &$0.43^*$\\
    WE, reductions &$0.88$&$0.035$& $0.59^{**}$&$0.06$\\
    WE, adversarial &$0.97$&$0.15$& $0.55^{**}$&$0.13$\\\midrule
    
    BERT, pre-process& $0.84$ & $0.11$ & $0.46^*$ & $0.08$\\
   BERT, post-process & $0.85$&$0$ & $0.46^*$&$-0.04$\\
   BERT, decoupled &$0.85$ & $0.22$&$0.48^*$&$-0.05$\\\bottomrule
\end{tabular}	

\end{small}	
\end{center}

\end{table}

\setcounter{footnote}{4}
\footnotetext{\label{pvalue}
We compute p-values for the two-sided test of zero correlation between $p_c$ and $r_c$ using SciPy’s \texttt{spearmanr} function \citep{virtanen2020scipy}. 
Values marked with $^*$ and $^{**}$ indicate that the p-value is $ < 0.05$ and $< 0.01$ respectively.}
\setcounter{footnote}{5}
\footnotetext{CoCL \citep{romanov2019s} is modulated by a hyperparameter $\lambda$ that determines the strength of the fairness constraint. We use $\lambda = 2$, which \citet{romanov2019s} finds to have the smallest \rmsgap on the occupation classification task. }
We find that the post-processing approaches have the largest value of \cov, i.e. the strongest associations with gender norms. 

Since post-processing does not change the ordering within a group, $r_c$ and \cov~remain identical to that of the approach without any fairness intervention (Figure \ref{fig:ppcorr}). Thus, post-processing continues to privilege individuals who align with the occupation's gender norms. Even when the desired group fairness metric is perfectly met, i.e. \rmsgap $= 0$, the \snob~correlations remain untouched. For post-processing, the group fairness and \snob~metrics are unrelated; the mitigation of one does not affect the presence of the other, which can be explained by the within-group order-preserving nature of post-processing approaches.

The pre- and in-processing approaches mitigate this observed association, as indicated by their lower \cov~compared to that of post-processing. Table \ref{tab:faircompare}
 reveals that unlike for post-processing, there are more complex relationships between \rmsgap~and \cov~for pre- and in-processing approaches. They generally improve both \rmsgap~and \cov. However, the improvements are not monotonic; for BOW, \rmsgap~is smaller for pre-processing while \cov~is larger compared to the decoupled approach in Table \ref{tab:faircompare}. The adversarial learning approach achieves the smallest value of \cov, although its \rmsgap~is the highest. This is an intuitive result, as it directly optimizes to reduce association between the target predictions and gender group membership. In-processing approaches are most effective at minimizing \snob, as shown in Table~\ref{tab:faircompare} and Figure~\ref{fig:approaches}. Notably, the in-processing techniques have access to features other than sensitive group membership and the target variable, and aim to penalize reliance on spurious correlations. 
 Appendix \ref{wordweight} provides further insight into the mechanisms behind why the classifiers may rely on gender norms. Since in-processing approaches are typically more expensive to implement, this highlights trade-offs between ease of implementation, classifier accuracy, and association with gender norms.

\section{Significance of Inferred Social Norms}\label{realworld}
\begin{figure}[h]
  \centering
  \includegraphics[width=\linewidth]{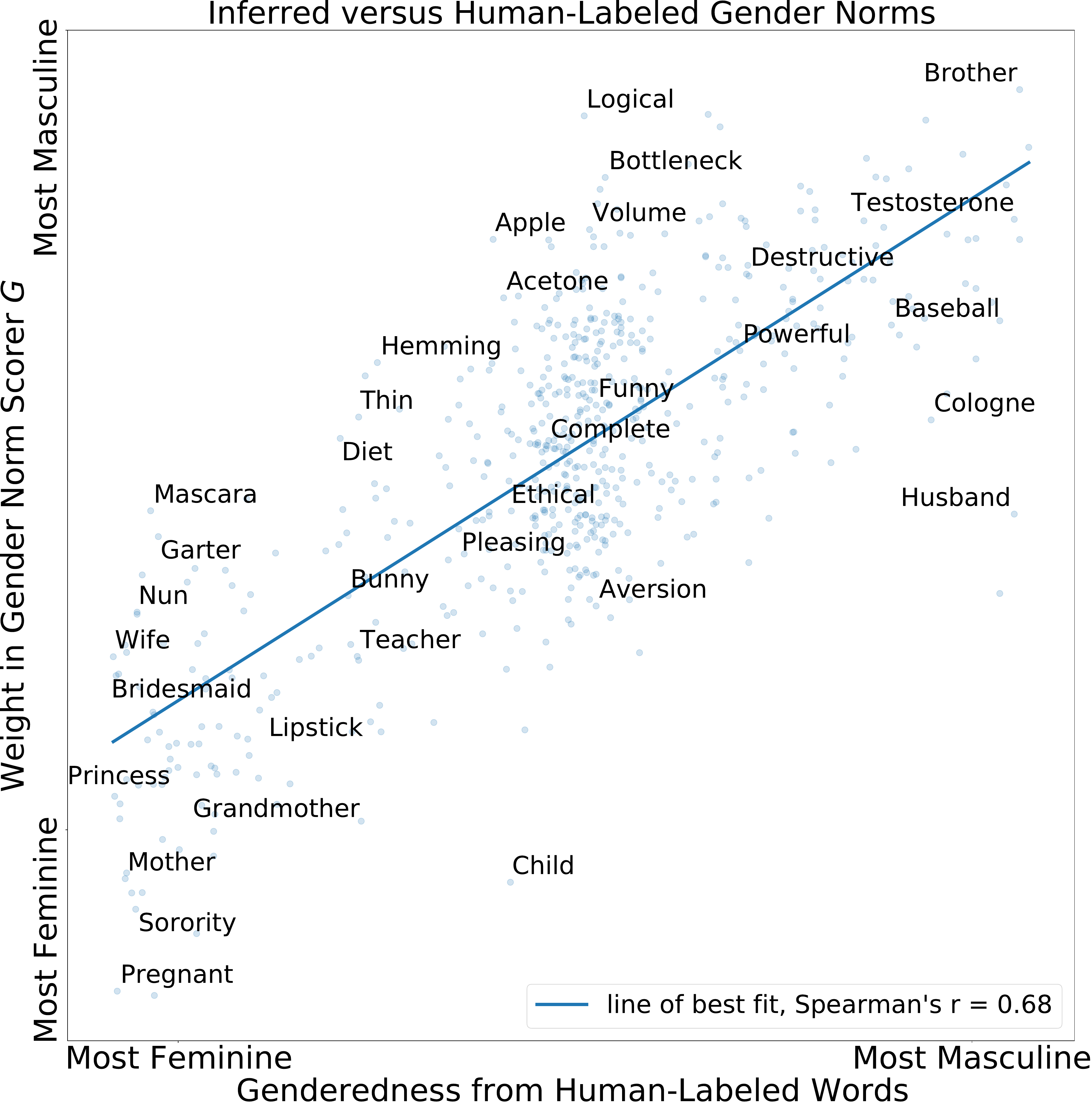}
  \caption{We use a human-labeled corpus to validate that the scorer $G$ infers gender norms that reflect a notion of real-world gender norms. Comparing the words' human-labeled gender scores ($x$-axis) to their weights in $G$ ($y$-axis), we find a strong correlation between the norms learned by $G$ and the human-labeled gender scores.}
    \label{fig:crawfordcorr}
\end{figure}

The potential harms of relying on inferred social norms for classification depend on what is captured by these inferred norms. To measure correspondence between inferred gender norms in the dataset and real-world gender norms, we use a human-labeled corpus of 600 words with gender scores labeled via crowdsourcing \citep{crawford_leynes_mayhorn_bink_2004}. For words in the corpus, we compare the human-labeled gender scores with their weights in the gender norm scorer $G$. The weight of a word $w$ in $G$ is $$\frac{e_w\cdot W_G}{|e_w||W_G|},$$ i.e. the cosine similarity between its embedding $e_w$ and the coefficient weight vector $W_G$ learned by $G$. The magnitude quantifies the word's importance to $G$, while the sign indicates the direction of the association. We find a strong correlation (Spearman's $r$-value 0.68) between these values (Figure \ref{fig:crawfordcorr}). In Section \ref{robust}, we perform a robustness test using a gender norm scorer that omits occupation-relevant words and find the same result. Thus, the inferred norms relate to existing social norms and can lead to compounding harms for individuals who do not adhere to these norms.

\subsection{Robustness of Associations: Occupation-Irrelevant Gender Norm Scorers}\label{robust}

\begin{figure}
  \centering
  \includegraphics[width=\linewidth]{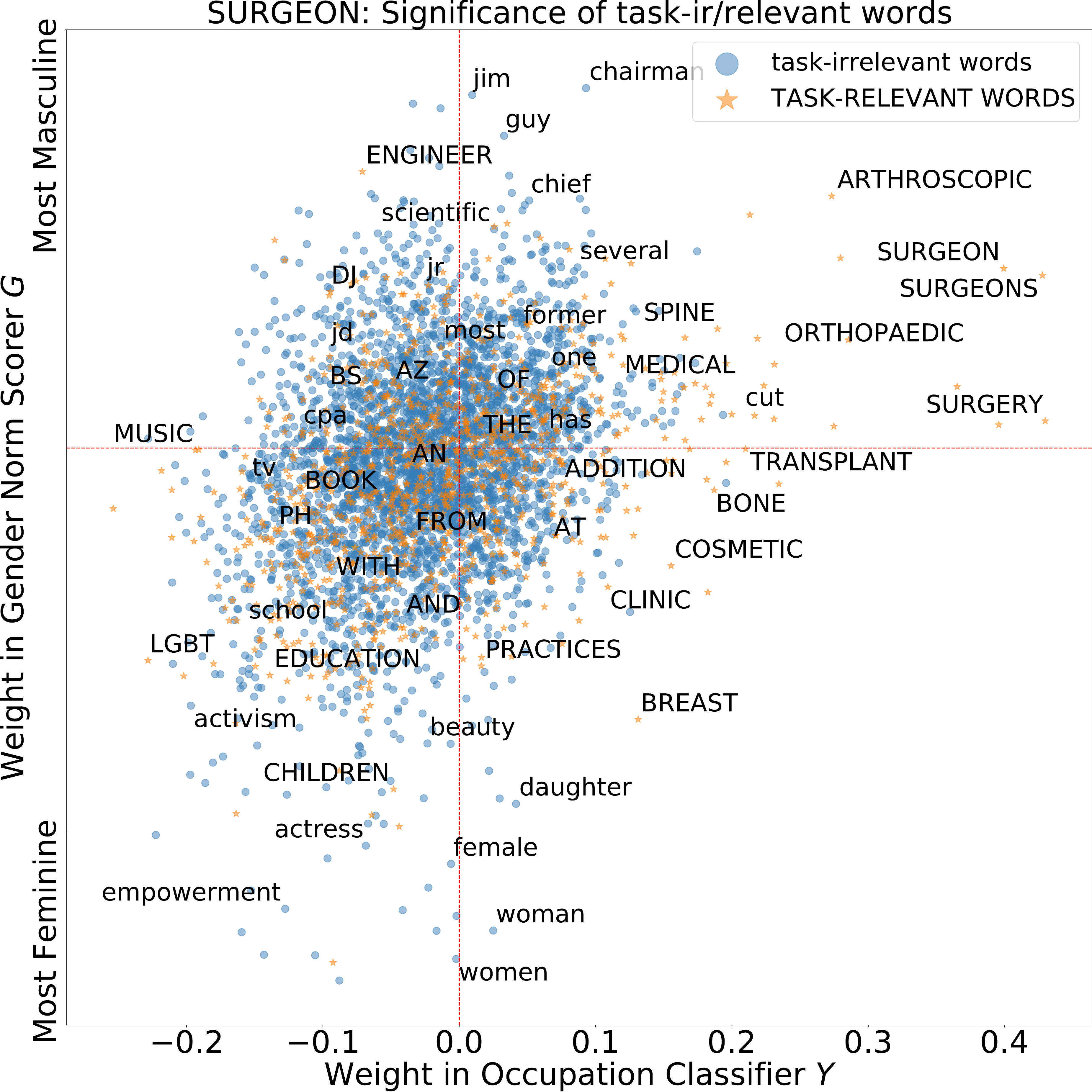}
  \caption{\textbf{Identifying task-irrelevant words.} Each word is plotted based on its weight in the occupation classifier ($x$-axis) and gender norm scorer ($y$-axis). Task-relevant and task-irrelevant words are labeled as uppercase and lowercase respectively. \girrev~uses only the task-irrelevant words to infer gender norms. Words highly predictive of the occupation (large $x$-value) are determined as task-relevant and thus not used in \girrev.  }
    \label{fig:words}
\end{figure}

For some occupations, certain ``gendered words'' may be acceptable for the occupation classifier to use and thus should not be considered as evidence of \snob. For example, the word “computer” is learned to be weakly associated with masculine gender norms by $G$ (``computer'' has weight 0.124 in the WE approach), but it does not seem discriminatory if the occupation classifier relies on ``computer'' to determine whether an individual is a software engineer. The word ``computer" is relevant to the task of determining this occupation. We define a word $w$ as task-relevant to occupation $c$ if it occurs significantly more frequently in the biographies of $c$ when conditioned on gender, i.e. $w$ occurs more in $S_c$ compared to all  other ``she" biographies in the dataset, and $w$ occurs more in $H_c$ compared to all other ``he" biographies in the dataset. Thus, it is reasonable that a statistical machine learning model uses $w$ as a predictor of $c.$ 

More broadly, one could worry that the \snob~identified using \cov~in Section~\ref{diff} could be attributed to the gender norm scorer using occupation, and task-relevant words, as a proxy. While we have addressed this concern by pre-processing the data to balance gender within each occupation, this section presents an additional robustness check. We train gender norm scorers \girrev~using only the words that are \textit{task-irrelevant} for occupation $c$.

To determine which words are task-ir/relevant, we measure the differences between 
\begin{enumerate}
    \item $\textsc{Freq}_w(S_{c'})$ versus $ \textsc{Freq}_w(\{S_c\}|c\in C),$
    \item $\textsc{Freq}_w(H_{c'})$ versus $ \textsc{Freq}_w(\{H_c\}|c\in C),$
\end{enumerate}
 where $\textsc{Freq}_w(S_{c'})$ denotes the number of times $w$ appears in $S_{c'}$, the ``she'' biographies of occupation $c'$. $\textsc{Freq}_w(H_{c'})$ is defined analogously.
 
We determine task-relevance conditioned on gender using a chi-squared test for each occupation and word. To account for multiple testing, we apply the Benjamini-Hochberg procedure \citep{benjamini1995controlling} to control the false discovery rate at the level 0.05.\footnote{We computed the p-values using the \texttt{fdrcorrection} method from the statsmodels Python package \citep{seabold2010statsmodels}.}

The false discovery rate reflects a trade-off between false positives and false negatives. If too many words are erroneously labeled as task-relevant, then the data will lack words for \girrev~to provide insight into adherence to gender norms. To verify that the scores from \girrev~relate to real-world social norms, we compared the vocabulary's weights to the human-labeled gender scores as in Section \ref{langgender}. Across the classifiers, they have average Spearman's correlation 0.58 with standard deviation 0.04.

On the other hand, if we erroneously label task-relevant words as task-irrelevant, then \girrev~may continue to reflect occupation-relevant words rather than a notion of gender norms. We choose an FDR that prioritizes mitigating this error and eliminating the use of task-relevant words by the gender norm scorer. Our method is cautious in determining ``task-irrelevant'' words: it may incorrectly label some words as task-relevant and exclude them from the norm scorer \girrev. 


Figure \ref{fig:words} illustrates the weights in the occupation classifier and gender norm scorer for task-irrelevant versus task-relevant words. We observe that the words with high weights in the occupation classifier are more likely to be labeled as task-relevant. Around 40\% of the words are labeled as task-irrelevant for each occupation. The resulting gender norm scorers, \girrev, have average accuracy 0.61 with standard deviation 0.04.

\begin{figure}
  \centering

      \includegraphics[width=\linewidth]{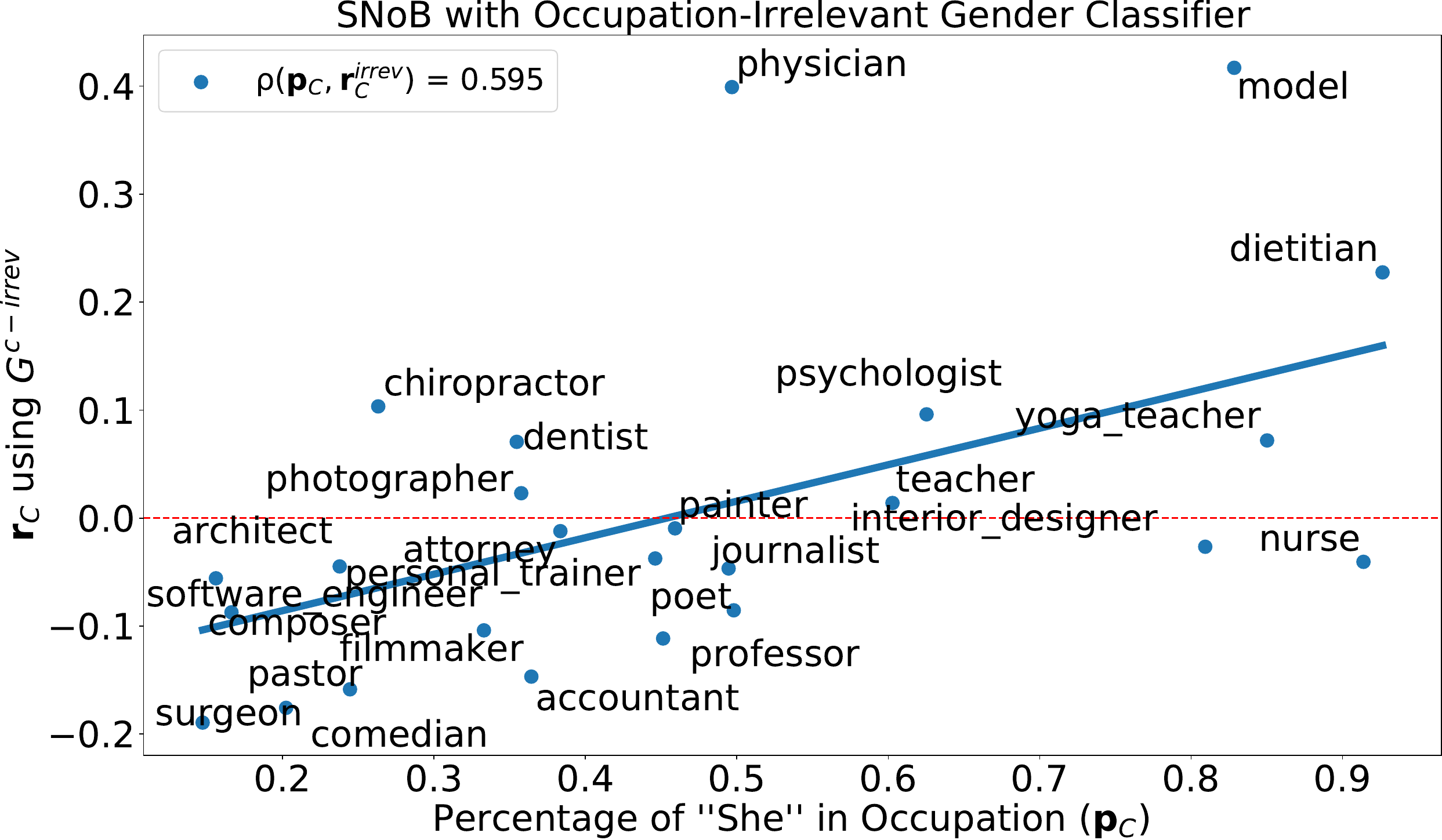}
    \vspace{0.3cm}

      \includegraphics[width=\linewidth]{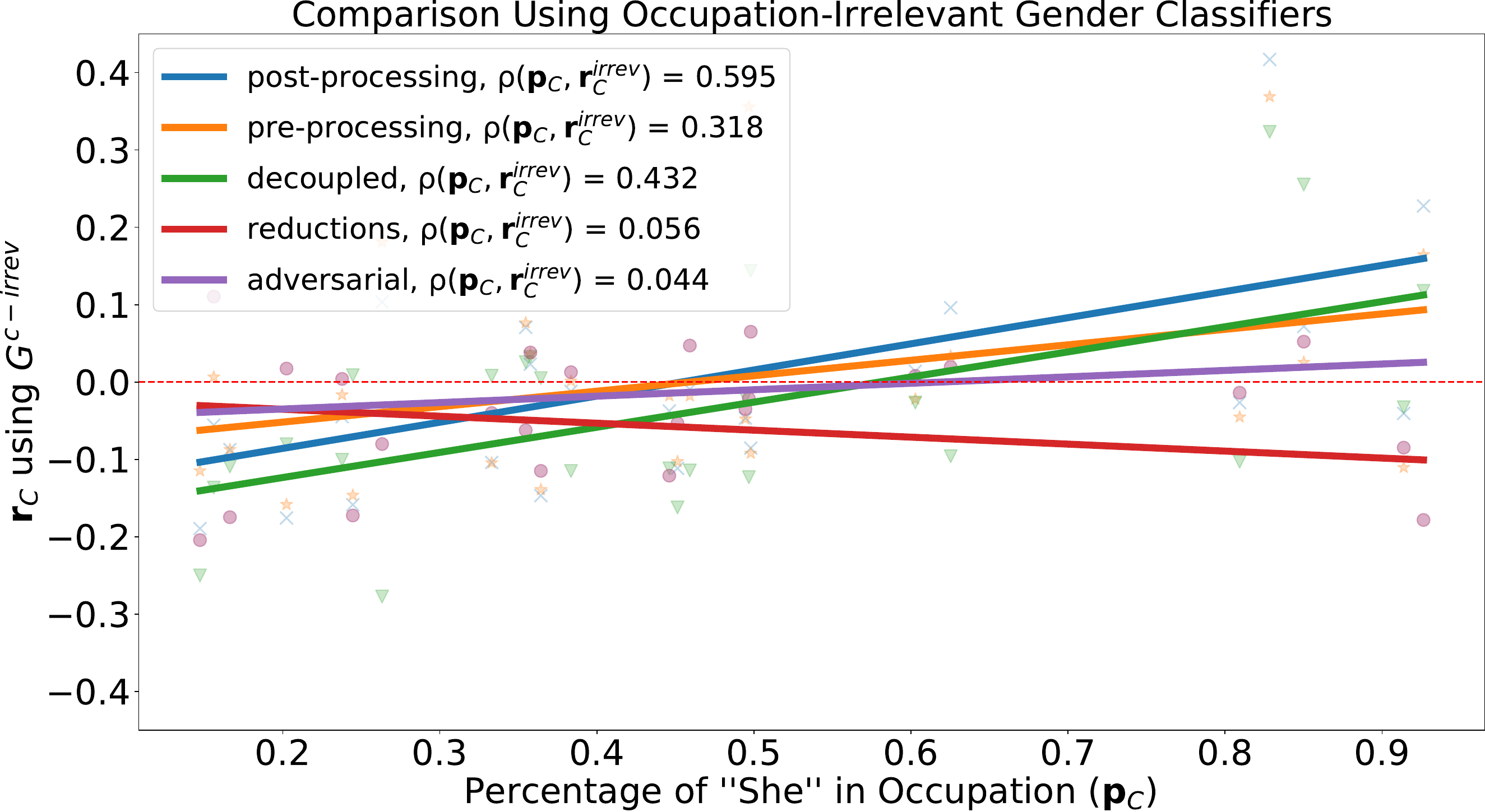}

  \caption{\textbf{\snob~using \girrev}. 
These plots measure the same associations as Figures \ref{fig:ppcorr} and \ref{fig:approaches}, using \girrev~instead of $G$ to obtain gender scores. The trends are similar, which suggests that our method effectively captures masculine and feminine \snob.
}
    \label{fig:irrevcorr}
\end{figure}

We compute \rirrev~and \covirrev, which are analogs of $r_c$ and \cov~ respectively, using the outputs obtained from \girrev~ in place of $G$. \covirrev~is listed in the last column of Table \ref{tab:faircompare}. Figure \ref{fig:irrevcorr} shows the trends of \rirrev~and \covirrev~ across occupations for different approaches. While the magnitude of \covirrev~is smaller than \cov, we find the same trends.
$r_c$ for male-dominated occupations are more negative, and vice versa for female-dominated occupations. Post-processing approaches have the highest values of \covirrev, while in-processing approaches most strongly mitigate this effect.
These results suggest that our proposed metrics capture meaningful notions of inferred gender norms beyond occupation-relevant words, thus effectively quantifying \snob.

\section{Conclusion and Future Work}
Our work introduces the risk of social norm bias, a type of harm that is not considered by group fairness techniques, and proposes a framework to quantify it.
We measure associations between algorithmic predictions and inferred gender norms in occupation classification, a task that is directly relevant to automated recruiting \citep{de2019bias, peng2019you}, and show that the norms inferred from the dataset correspond to real-world gender norms. We reveal that \snob~is the strongest in post-processing approaches. The differences found across pre-, post-, and in-processing approaches should be considered when choosing between these approaches. In particular, our results introduce considerations that may justify the costs of re-training an algorithm rather than relying on post-processing approaches.

More broadly, we characterize how algorithms discriminate based on \snob, a non-categorical aspect of a sensitive attribute. 
Our approach can be generalized to capture \snob~in other types of data that reflect social norms, such as images, behavioral data, and other forms of structured data.
By illuminating a new axis along which algorithmic discrimination occurs, our work shows a type of harm that stems from the reliance of ``group fairness'' approaches on reductive definitions. Furthermore, while our work has centered the harm of \emph{deviating} from norms associated with a majority, there may also be a risk of harm from \emph{adhering} to norms associated with a previously marginalized group, which future work should explore. For example, if an algorithm systematically overestimates risk for members of a group (e.g. risk of loan default), it is possible that fairness-aware algorithms that equalize a group fairness metric still penalize those who adhere to certain norms of the group. 

\snob~may be exhibited on multiple axes of sensitive attributes, such as social norms related to race, class, and gender. Based on real-world concerns of how intersectional identities can compound bias \citep{crenshaw1990mapping}, the concept of \snob serves a starting point to characterize how the intersections among these different dimensions may affect an algorithm. In future work we intend to explore algorithmic approaches to reducing these associations.

\paragraph{Limitations}\label{lim}
As \citet{wojcik_remy_2020} note, an algorithm's notion of gender is based only on observable characteristics. There are many other elements of social norms that our methods do not capture. The broader notions of gender expression and gender identity are ever-changing cultural phenomena  \citep{scheuerman_paul_brubaker_2019}. Also, eliminating \snob~does not necessarily remove other sources of bias.
\bibliographystyle{ACM-Reference-Format}
\bibliography{references}
\begin{table*}[h]
  \caption{Correlation \rprofnb~ (first three columns) and \rprof~ (latter three columns) across pre-processing (pre-proc), post-processing (post-proc), and decoupled approaches. For DE on the nonbinary dataset, we consider both classifiers, trained on  the ``she'' and ``he'' data respectively.\protect\footnotemark{}
  }
  \label{tab:nb}
\begin{center}
\begin{small}
\begin{tabular}{l|ccc|ccc}
\toprule
Correlation&\multicolumn{3}{c|}{\rprofnb}&\multicolumn{3}{c}{\rprof}\\\midrule
Approach&pre-process&post-process&decoupled&pre-process&post-process&decoupled\\\midrule
BOW& $0.36$ &
$0.34$ &
$0.35, 0.32$ &$0.09^{**}$&$0.10^{**}$&$0.11^{**}$\\
WE& $0.29$ &$0.29$& $0.31, 0.31$&$-0.01^{**}$&$-0.02^{**}$&$-0.01^{**}$\\\bottomrule
\end{tabular}	
\end{small}	
\end{center}	

\end{table*}
\pagebreak
\appendix

\section{Gendered Words Used in Classifiers}\label{wordweight}
We provide insight into some of the differences across the classifiers that may be driving the \snob~ described in preceding sections. We define $\beta_w$ as  the weight of a word $w$ based on the value of the classifiers' coefficients. We focus on the logistic regression classifiers using the BOW and WE representations since the BERT representations are contextualized, so each word does not have a fixed weight to the model that is easily interpretable.

For the BOW representation of a biography $x$, each feature in the input vector $v_x$ corresponds to a word $w$ in the vocabulary. We define $\beta_w$ as the value of the corresponding coefficient in the logistic regression classifier. The magnitude of $\beta_w$ is a measure of the importance of $w$ to the occupation classification, while the sign (positive or negative) of $\beta$ indicates whether $w$ is correlated or anti-correlated with the positive class of the classifier.

For the WE representation, we compute the weight of each word as $$\beta_w = \frac{e_w\cdot W_c}{|e_w||W_c|},$$ i.e. the cosine similarity between each word's fastText word embedding $e_w$ and the coefficient weight vector $W_c$ of the WE-representation classifier. Like in the BOW representation, the magnitude of $\beta_w$ quantifies the word's importance, while the sign indicates the direction of the association.

If a word $w$ has positive/negative weight for classifier $Y_c$, then adding $w$ to a biography $x$ increases/decreases the predicted probability $Y_c(x)$ respectively.

Let $\beta_w(Y_c)$ be the weights for approach $Y_c.$ We examine the words whose weights $\beta_w$ satisfy
\begin{enumerate}
\item $|\beta_w(Y_c)| > T$,
    \item $|\beta_w(G)| > T$,
    \item $|\beta_w(Y_c)| > T'\cdot|\beta_w(Y_c')|$,
\end{enumerate}
where $T, T'$ are significance thresholds and $Y_c, Y_c'$ are two different occupation classification approaches.

Words that satisfy these conditions are not only associated with either masculinity or femininity but also weighted more highly in approach $Y_c$ compared to $Y_c'$. Thus, including these gendered words in a biography influences $Y_c$'s classification more strongly than that of $Y_c'$. This suggests that they may contribute more strongly to the \cov~in one approach than the other. 
For example, we examined these words for the occupations of \textsc{surgeon, software engineer, composer, nurse, dietitian}, and  \textsc{yoga teacher}, which are the six most gender-imbalanced occupations, with $Y_c = $ \{BOW, post-processing\}, $Y_c' =$ \{BOW, decoupled\}, $T = 0.5$ and $T' = 0.7$. The words that satisfy these conditions are ``miss'', ``mom'', ``wife'', ``mother'', and ``husband." Conversely, with $Y_c = $ \{BOW, DE\} and $Y_c' =$ \{BOW, PO\}, the words are ``girls'', ``women'', ``gender'', ``loves'', ``mother'', ``romance'', ``daughter'', ``sister'', and ``female.''
These gendered words illustrate the multiplicity of gender present in the biographies beyond categorical labels, which standard group fairness interventions do not consider. 

Our analysis is limited by the fact that we only consider the individual influence of each word conditioned on the remaining words, while the joint influence of two or more words may also be of relevance.

\section{Analysis on Nonbinary Dataset}\label{nbanal}

\footnotetext{The p-value corresponding to \rprofnb~is for the hypothesis test whose null hypothesis is that the rankings from $G$ and $Y$ are uncorrelated. Values marked with a * indicate that the p-value is $< 0.05.$ Values marked with ** denotes that the p-value is $< 0.01.$}
We aim to consider how algorithmic fairness approaches affect nonbinary individuals, who are overlooked by group fairness approaches \citep{keyes2021you}. Using the same regular expression as~\citet{de2019bias} to identify biography-format strings, we collected a dataset of biographies that use nonbinary pronouns such as ``they'', ``xe'', and ``hir.'' Since ``they'' frequently refers to plural people, we manually inspected a sample of 2000 biographies using``they'' to identify those biographies that refer to individuals. \textsc{professor} is the only occupation title with more than 20 such biographies; the other occupations have too few biographies to perform meaningful statistical analysis. We computed \rprofnb, which is analogous to \rprof, the measure of \snob~for an individual occupation classifier introduced in Section \ref{singleocc}. While \rprof~is Spearman's correlation computed across the biographies in $S_c$, \rprofnb~ is the correlation across the nonbinary biographies in the profession.
The results are reported in Table \ref{tab:nb}. We find that \rprofnb~is positive across different approaches. However, the associated p-values are quite large $(>0.1)$, so it is challenging to analyze these associations. This is likely due to the small sample size; while \rprof~ is computed across the 10677 professor biographies that use ``she'' pronouns, \rprofnb~ is across only 21 biographies.

\begin{figure*}
  \centering
    \includegraphics[width=0.53\linewidth]{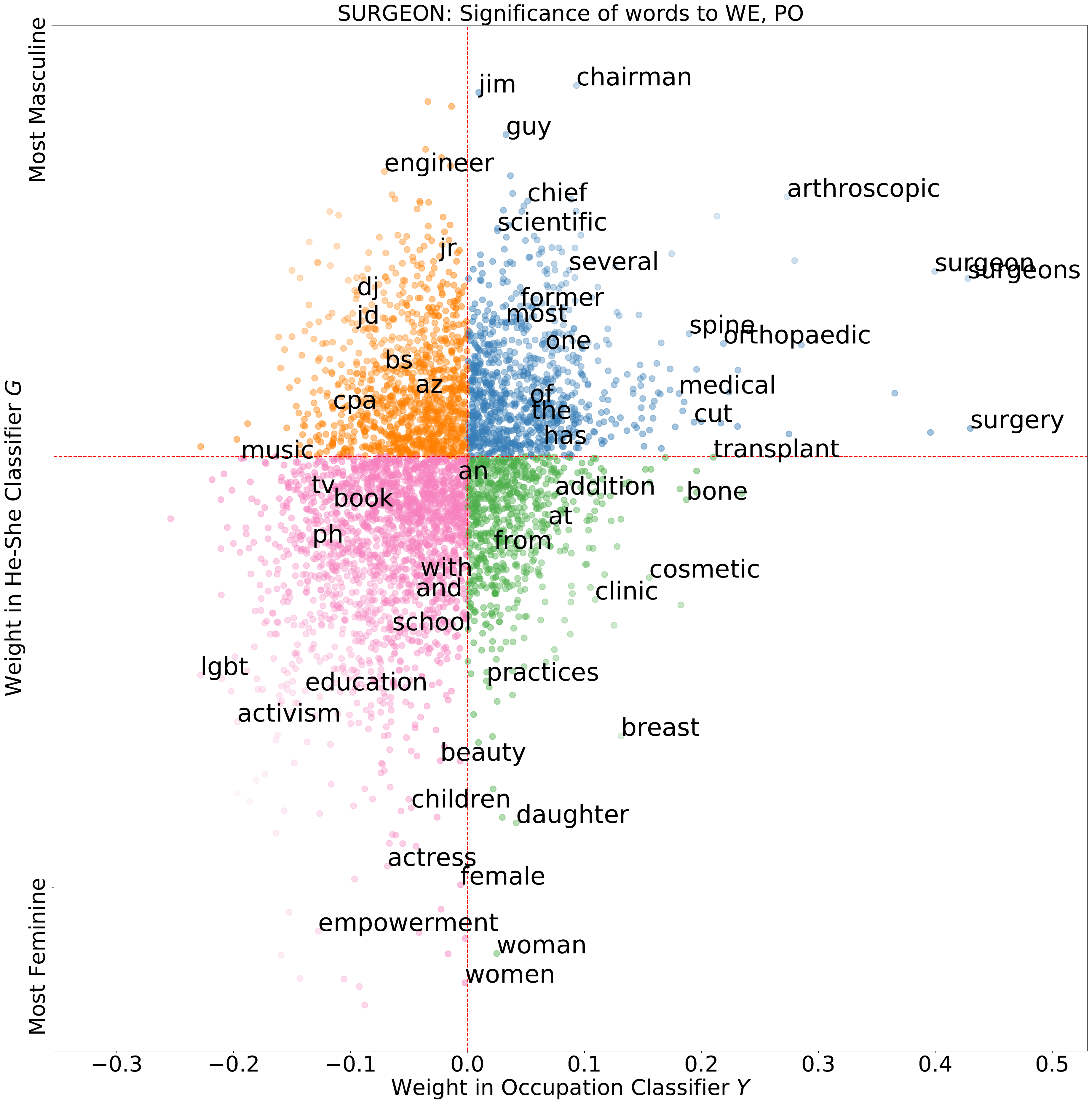}
  \includegraphics[width=0.53\linewidth]{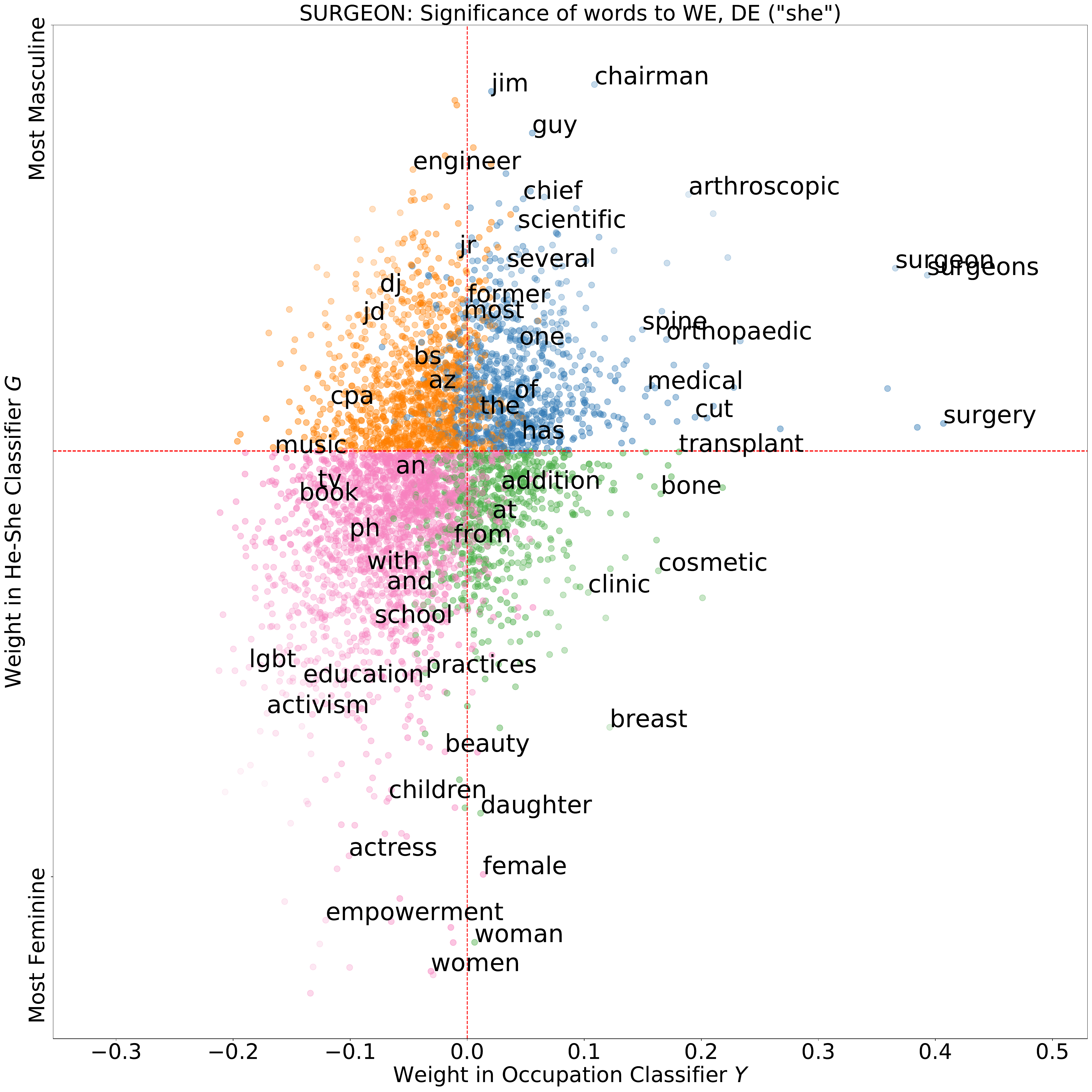}
  \caption{Words' weights in the occupation and gender classifiers for different approaches in the \textsc{surgeon} occupation. Each point represents a word; its $x$-position and $y$-position represents its weight in $\hat{Y}_c$ and $G$ respectively. Each point is colored based on its quadrant in the post-processing approach. Many points are closer to the $y-$axis in the decoupled approach. }
    \label{fig:weights}
\end{figure*}

\section{Word Weights}

In Figure \ref{fig:weights}, we plot the weight of each word in the BOW vocabulary in the occupation classifiers and gender classifiers.  These weights illuminate some of the mechanisms behind the predictions. Ideally, without \snob, every point would have small magnitude in either the occupation or gender classifier, i.e. lie on either the $x-$ or $y-$axis of Figure \ref{fig:weights}. We observe that in the DE approach, words are closer to the $y-$axis compared to the post-processing approach. This corresponds to the smaller value of \cov~exhibited by the decoupled approach compared to the post-processing one in Table \ref{tab:faircompare}.
Note that the post-processed classifier is trained on all of the biographies, while the decoupled classifier is trained on only biographies that use the same pronoun.

\end{document}